\documentclass[letterpaper, 10 pt, conference]{ieeeconf}  

\IEEEoverridecommandlockouts                              

\usepackage{times}
\usepackage{amssymb}
\usepackage{pifont}
\newcommand{\cmark}{\ding{51}}%
\newcommand{\xmark}{\ding{55}}%

\usepackage{algorithm}
\usepackage[noend]{algcompatible}

\usepackage{amsmath}
\usepackage{amsfonts}
\usepackage{breqn}
\usepackage{multirow,hhline,graphicx,array}
\graphicspath{ {./images/} }
\usepackage{multicol}
\usepackage[bookmarks=true]{hyperref}
\usepackage{subfiles}
\usepackage{caption}
\usepackage{subcaption}
\usepackage{xcolor}
\usepackage{footnote}
\usepackage[flushleft]{threeparttable}

\title{\LARGE \bf
Contrastive Learning of Features between Images and LiDAR
}

\author{Peng Jiang$^{1}$ and Srikanth Saripalli$^{1}$
\thanks{$^{1}$Peng Jiang ({\tt\small maskjp@tamu.edu}) and Srikanth Saripalli ({\tt\small  ssaripalli@tamu.edu}) are with the J. Mike Walker '66 Department of Mechanical Engineering, Texas A\&M University,
College Station, TX 77840, USA}%
}

\begin{document}

\maketitle
\thispagestyle{empty}
\pagestyle{empty}

\begin{abstract}
Image and Point Clouds provide different information for robots. Finding the correspondences between data from different sensors is crucial for various tasks such as localization, mapping, and navigation. Learning-based descriptors have been developed for single sensors; there is little work on cross-modal features. This work treats learning cross-modal features as a dense contrastive learning problem. We propose a  Tuple-Circle loss function for cross-modality feature learning.
Furthermore, to learn good features and not lose generality, we developed a variant of widely used PointNet++ architecture for point cloud and U-Net CNN architecture for images. Moreover, we conduct experiments on a real-world dataset to show the effectiveness of our loss function and network structure. We show that our models indeed learn information from both images as well as LiDAR by visualizing the features. 
\end{abstract}
\section{INTRODUCTION}
Camera and Light Detection and Ranging (LiDAR) sensors are essential components of autonomous vehicles with complementary properties. A camera can provide high-resolution color information but is sensitive to illumination and lacks direct spatial measurement. A LiDAR can provide accurate spatial information at long ranges and is robust to illumination changes, but its resolution is much lower than the camera and does not measure color. However, combining the information from two sensors is always difficult due to the differences in representation (pixel vs. point) and information (visual vs. geometric), etc. Common methods combine the two sensors' information by manually defining common features across two modalities like edges, gradients, and semantics. This is widely used in traditional calibration methods \cite{Jiang2021_MFI}, which is always the first step when using multiple sensors. Other scenarios include mapping, localization, SLAM etc. \cite{Pham2019_LCD, Wang2022_p2net, Feng2019_2D3Dmatchnet}. Thus, defining common features across sensor modalities is important across all applications.

With the development of deep learning, many methods have been explored in the literature to generate features for a single modality like image \cite{Wang2021_DenseCL} or point cloud \cite{Christopher2019_ICCV}. The features can be used to find the inter-modal correspondences but cannot directly be applied to cross-modal situations. Finding unified cross-modal dense features is more challenging than finding unified inter-modal features. A few works have worked on cross-modal descriptors, and most of them are patch-based methods and require dense point clouds to learn\cite{Pham2019_LCD, Wang2022_p2net, Feng2019_2D3Dmatchnet}. One reason is that there are no unified neural network structures that are versatile to process all modalities. Images are always processed by 2D convolutional networks.
Meanwhile, various architectures can process point clouds according to different representations. This difference makes it difficult to train two models simultaneously. Secondly, compared with other 3D scanners, LiDAR creates sparser point clouds, making this problem much more challenging. Moreover, \cite{Wang2022_p2net} and our preliminary tests show that using the existing loss functions is challenging for point clouds and image cross-feature learning.

To overcome the above difficulties but not lose generality, we propose a U-Net variant of the widely used PointNet++\cite{Qi2017_NIPS} architecture for the point cloud and a U-Net CNN architecture for images. To handle the sparsity of LiDAR, we add multi-scale grouping features to the feature propagation layer of PointNet/PointNet++. We propose a Tuple-Circle loss function by viewing the feature learning problem from a contrastive learning perspective and considering the cross-modal properties to optimize the two networks. The contributions of this paper are summarized as follows:
\begin{itemize}
  \item We propose the Tuple-Circle loss function for cross-modal deep contrastive learning. By representing the features vectors into tuples and adding self-paced weights, Tuple-Circle loss helps the models learn features across different  modalities
  \item We propose U-Net architecture for 2D image and 3D point cloud dense feature contrastive learning with flexible receptive fields.
  \item We conduct experiments on a real-world dataset KITTI360\cite{Liao2021_ARXIV} to show the effectiveness of our method and visualize the learned features. 
\end{itemize}
\section{RELATED WORK}
\subsection{Contrastive Learning}
Contrastive representation learning aims to learn a feature space where similar sample pairs are close to each other and dissimilar pairs are away. Several contrastive learning loss functions have proposed \cite{Le2020_ACCESS, Sun2020_Circleloss}.  \cite{Chopra2005_CVPR} is one of the earliest works about contrastive learning. It proposes Contrastive loss that considers only one positive or negative pair at a time. The triple loss \cite{Schroff2015_CVPR} tries to minimize the similarity between anchor samples and positive samples and simultaneously maximize the similarity between anchor samples and negative samples. \cite{Sohn2016_NIPS} generalizes triple loss to include comparison with multiple negative samples. \cite{Gutmann2010_PMLR} proposes NCE(Noise Contrastive Estimation) by using logistic regression to recognize positive samples from one noise (negative samples). Inspired by NCE, \cite{Van2018_arXiv} proposes  InfoNCE loss which uses categorical cross-entropy loss to deal with multiple negative samples.
Circle loss in \cite{Sun2020_Circleloss} provides a uniﬁed perspective for optimizing pair similarity. For multimodal contrastive learning,  \cite{Liu2021_ICCV} uses tuples to differentiate information from different modalities and comes up with TupleInfoNCE loss for multimodal fusion task. Most of the above methods are used for global embedding learning. Some methods attempt to learn the dense embedding of inputs. 
\cite{Wang2021_DenseCL} implements a self-supervised learning method by optimizing a pairwise contrastive (dis)similarity loss at the pixel level between two views of input images. \cite{Christopher2019_ICCV} proposes hardest-contrastive losses and hardest-triplet losses by exploring hard negative samples to learn geometric features of the point cloud. 
Our Tuple-circle loss considers the cross-modal properties of cross-modal learning and provides a more flexible and adaptive property for optimization.  
\subsection{Image and Point Cloud Feature Learning}
\cite{Cattaneo2020_ICRA} uses the triple loss and combines the teacher/student method to create a shared 2D-3D embedding space for image-based global localization in LiDAR-maps. \cite{Yin2021_i3dLoc} utilizes a Generative Adversarial Network (GAN) to extract cross-domain symmetric place descriptors for localizing a single camera with respect to a point cloud map for indoor and outdoor scenes. \cite{Xing2018_3DNet} is designed for learning robust local feature representation leveraging both textures from images and geometric information from the point cloud.
\cite{Li2015_JointEmbeddings} uses CNN to map an image to a point in the embedding space, which is created by using a 3D shape similarity measure. The embedding allows cross-view image retrieval, image-based shape retrieval, as well as shape-based image retrieval. In \cite{Feng2019_2D3Dmatchnet} an end-to-end deep network architecture is presented to jointly learn the descriptors for 2D and 3D keypoints from image and point cloud, respectively. As a result, the approach is able to directly match and establish 2D3D correspondences from the query image and 3D point cloud reference map for visual pose estimation. \cite{Pham2019_LCD} proposes LCD that uses a dual auto-encoder neural network and triplet loss to learn a shared latent space representing the 2D and 3D data. Still, the method requires a point cloud with RGB information. 

Our approach is closest to \cite{Wang2022_p2net}. \cite{Wang2022_p2net} proposes a joint learning framework with an ultra-wide reception mechanism for simultaneous 2D and 3D local features description and detection to achieve direct pixel and point matching. Based on Circle loss, \cite{Wang2022_p2net} comes up with circle-guided descriptor loss to train P2-Net for joint description local features for pixel and point matching. However, circle-guided descriptor loss only considers cross-modality positive sample pairs and inter-modal negative sample pairs, which do not fully utilize batch data. It converges slowly in cross-modality sitting and fails in LiDAR and image situations. The model proposed in this paper, combined with the Tuple-Circle loss function, can work in LiDAR and image pairs situations.
\section{OUR APPROACH}
\begin{figure}[t]
\includegraphics[width=0.45\textwidth]{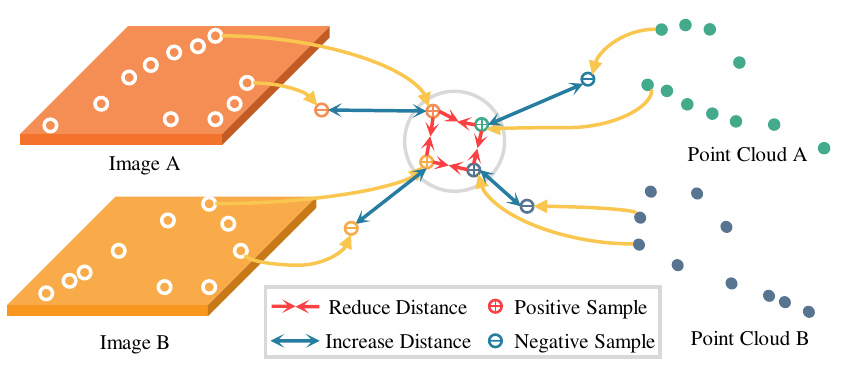}
\caption{\textbf{Cross-modal Contrastive Learning}: Image A/B and Point Cloud A/B are acquired from the same scene. The contrastive learning process is to try to reduce the distance between the positive pairs and increase the distance between negative pairs.} 
\label{fig:crossmodal_learning}
\end{figure}
\subsection{Tuple-Circle Loss}
Contrastive learning, an approach in self-supervised learning, allows the model to learn rich representative features. Contrastive representation learning aims to learn such a feature space where similar sample pairs stay close while dissimilar ones (called negative samples $x^{-}$) are far apart see Fig.\ref{fig:crossmodal_learning}. In a single modality setting, each sample in a dataset is treated as a different instant (called anchor sample $x$). Similar counterparts (called positive samples $x^{+}$) for training are created by random augmentations, e.g., rotating an image, gaussian blur. We consider an image-lidar pair acquired from the same scene and their augmented counterparts as a dataset in our setting and the pixel and point corresponding to the same physical locations as positive pair. 

It is ideal to have a loss function to deal with multiple positive and negative samples in a cross-modal learning setting because samples from different modalities can construct different pairs. Many loss functions have been proposed for contrastive learning. Sun et al. \cite{Sun2020_Circleloss} proposed a unified loss function from a pair similarity optimization viewpoint on deep feature learning. Based on the unified loss function, he came up with Circle Loss (see Eq.\eqref{eq:orig_circle_loss}) by adding self-paced weights and margins.
{\small 
\begin{equation}
\begin{aligned}
\mathcal{L}_{circle} &=\log \left[1+\mathcal{S}^{-}\mathcal{S}^{+}\right]\\
\mathcal{S}^{-} &=\sum_{j=1}^{L} \exp \left( \alpha_{j}^{-}\left(s_{j}^{-}-\bigtriangleup^{-}\right)\right)\\
\mathcal{S}^{+} &=\sum_{i=1}^{K} \exp \left(\alpha_{i}^{+}\left(s_{i}^{+}-\bigtriangleup^{+}\right)\right)\\
\end{aligned}
\label{eq:orig_circle_loss}
\end{equation}}
where 
$\alpha_{j}^{-}=\gamma\max(s_j^-+m,0)$ and $\alpha_{i}^{+}=\gamma\max(1+m-s_i^+,0)$ are self-paced weights. $L$ and $K$ are the total numbers of the neigative samples and positive samples. $\bigtriangleup^{-}=m$ and $\bigtriangleup^{+}=1-m$  are the negative pair and positive margin. There are only three hyper-parameters of circle loss in which $\gamma$ is a scale factor, $m$ controls the radius of the decision boundary. $s^{+}=f\left(x,x^{+}\right)$ represents similarity between features of positive samples, and  $s^{-}=f\left(x,x^{-}\right)$ denotes similarity between features of negative samples. 

\begin{figure*}[t!]
    \centering
    \includegraphics[width=\textwidth]{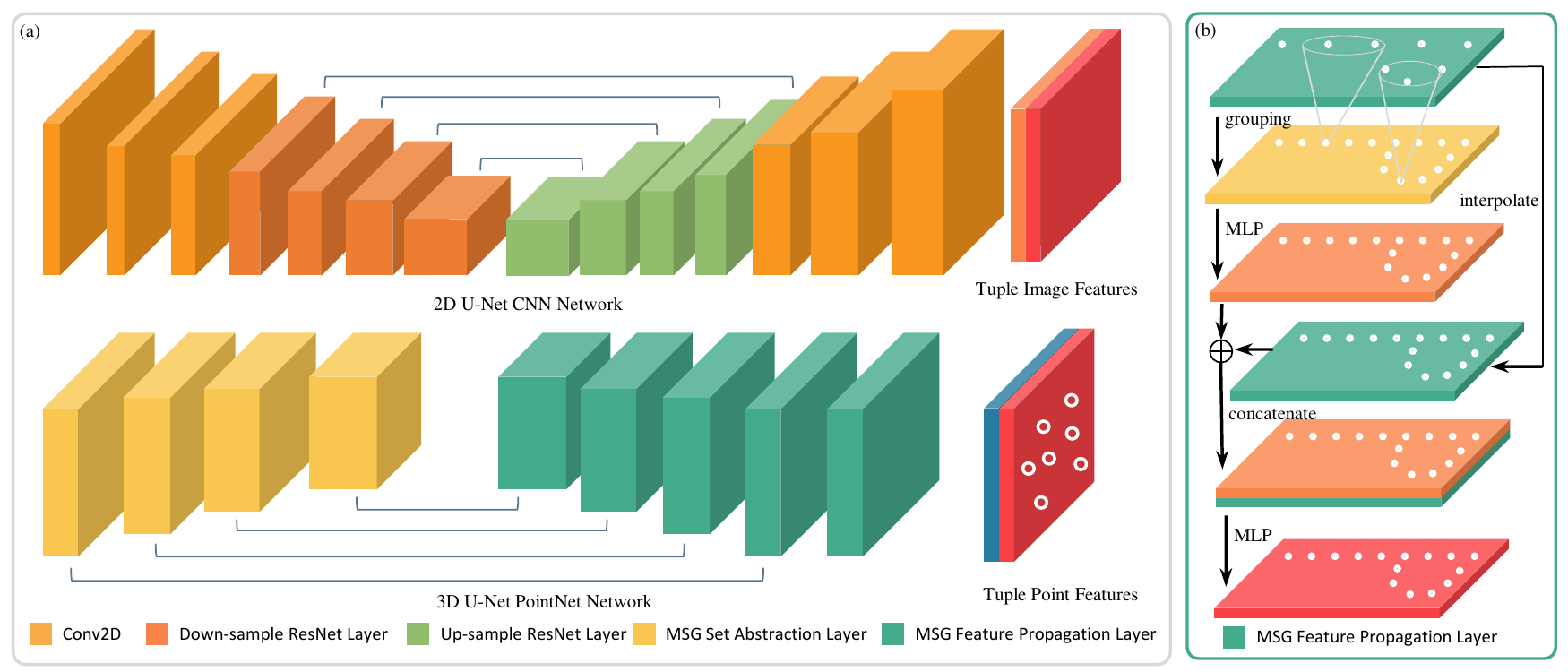}
    \caption{\textbf{Architecture of 2D and 3D Model} (a) The 2D model includes 3 Convolutional networks and four ResNet layers to downsample the input and 4 upsample the ResNet layer to upsample the features into the original size. The 3D model includes four multi-scale grouping set abstraction layers to downsample the input points and four multi-scale grouping feature propagation layers to upsample the features. (b) The modified feature propagation layers include multi-scale grouping operators.}
    \label{fig:framework}
\end{figure*}

Our preliminary test shows that self-pace weight plays a vital role in the convergence of cross-modality learning. The weights of negative pairs $\alpha_{j}^{-}$ decrease while the distance between the negative pairs increases. This behavior guarantees that the features from the two different modalities will not be pushed too far between each other before the two models converge to the same features space. The self-paced weights of positive samples $\alpha_{i}^{+}$ also have similar effects. As the two positive samples become close to each in the features, the weights decrease to allow optimization to take care of other pairs. 

However, similar to other approaches\cite{Chopra2005_CVPR, Sohn2016_NIPS, Van2018_arXiv}, circle loss also does not consider the properties of cross-modality learning and uses a unified representation to represent the features (same-sized vectors). We propose Tuple-Circle loss to generalize the circle loss to cross-modality feature learning. In a cross-modality setting, samples from different modalities can share some common information but may also contain modality-specific information \cite{Liu2021_ICCV, Jiang2021_LiDARNet, Bousmalis2016_NIPS}. Separate treatment of the common and modality-specific information can make learning easier. 

To learn features across K modalities, we represent the features as a k-tuple to separate the information $x_{f}=\left[x_{t1},x_{t2},\dots x_{tk}\right]$ in which $k-1$ features vectors are shared with other modality and $1$ vector is for private feature learning. For inter similarity measurement $s^{+}_{inter}=f\left(x,x^{+}\right)$, for cross-modality similarity, we have $s^{+}_{crossk}=f\left(x_{tk},x_{tk}^{+}\right)$. Therefore, our final loss function can be represented as:
{\small
\begin{equation}
\mathcal{L}_{tuple} =\log \left[1+(\mathcal{S}^{-}_{inert}+\mathcal{S}^{-}_{cross})(\mathcal{S}^{+}_{inert}+\mathcal{S}^{+}_{cross})\right]
\label{eq:tuple_loss}
\end{equation}}
where $S_{inter}^{+} and S_{inter}^{-}$ are sum of similarity measurement within modality as described in Eq. \eqref{eq:inter_measure} and $S_{cross}^{+} and S_{cross}^{-}$ sum of similarity measurement for cross modality as shown in Eq. \eqref{eq:cross_measure}.

In an Image-Point Cloud cross modality setting, we represent the features as $x_{f}=\left[x_{sh},x_{pr}\right]$ where $x_{sh}$ denotes shared features between image and point cloud and $x_{pr}$ represents modality-specific private features in point cloud or image. During training, we use one set of a positive pair$(x,x^+)$ and $N-1$ negative pairs $\left[(x,x^-_1),\dots,(x,x^-_{N-1})\right]$ from one modality. For inter-modal learning, we have:
{\small
\begin{equation}
\begin{aligned}
\mathcal{S}^{-}_{inter} =&\sum_{j=1}^{N-1} \exp \left(\alpha_{img_j}^{-}\left(s_{img_j}^{-}-\bigtriangleup^{-}\right)\right)\\
                 &+\sum_{j=1}^{N-1} \exp \left(\alpha_{pcd_j}^{-}\left(s_{pcd_j}^{-}-\bigtriangleup^{+}\right)\right)\\
\mathcal{S}^{+}_{inter} =& \exp \left( \alpha_{img}^{+}\left(s_{img}^{+}-\bigtriangleup^{+}\right)\right)\\
                  &+\exp \left( \alpha_{pcd}^{+}\left(s_{pcd}^{+}-\bigtriangleup^{+}\right)\right)\\
\end{aligned}
\label{eq:inter_measure}
\end{equation}}
For cross modality, we can have $4$ cross-modality positive pairs and $2N-2$ negative pairs by combination:
{\small
\begin{equation}
\begin{aligned}
\mathcal{S}^{-}_{cross} &=\sum_{j=1}^{2N-1} \exp \left(\alpha_{sh_j}^{-}\left(s_{sh_j}^{-}-\bigtriangleup^{-}\right)\right)\\
\mathcal{S}^{+}_{cross} &=\sum_{i=1}^{4} \exp \left( \alpha_{sh_i}^{+}\left(s_{sh_i}^{+}-\bigtriangleup^{+}\right)\right)\\
\end{aligned}
\label{eq:cross_measure}
\end{equation}}
In this paper, we use cosine similarity to measure the similarity between two features. For inter modality, we use the whole vector to compute cosine similarity $s^{+}=\frac{x_{f} x_{f}^{+}}{\vert x_{f} \vert\vert x_{f}^{+} \vert}$ and for cross modalities, we only use the shared part of the feature vector to compute the similarity $s^{+}_{sh}=\frac{x_{sh}x^{+}_{sh}}{\vert x_{sh} \vert\vert x^{+}_{sh} \vert}$.

\subsection{Models}\label{models}
We develop a dual U-Net structure framework(see Fig.\ref{fig:framework}) for cross-modality dense feature learning and describe the network structure next.

There are several approaches to learning from LiDAR data. One way is to project LiDAR data onto a spherical plane and process further using the 2D CNN network. By using this point cloud representation, we are able to unify both models under 2D CNN architecture\cite{Goodfellow2016_DeepLearning}. However, the projection process leads to the loss of 3D information of the point cloud. For generality of our model and allow our model to be applied to point clouds from other 3D sensors, we chose  PointNet/PointNet++ as the model for point cloud feature learning. PointNet/PointNet++ was designed to process raw point clouds directly, and several architectures for 3D data\cite{Zhang2019_Access} are designed on a similar structure and applied to our method. 

We first explore the vanilla multi-scale group (MSG) version of PointNet++, which was designed for the Point Cloud semantic segmentation task \cite{Qi2017_NIPS}. The model has an encoder consisting of a farest-sampling layer and sets abstraction layers, and the decoder is comprised of feature propagation layers. However, the model has difficulty producing good point-level features. A potential reason for this is that the feature propagation layers in the decoder create point features by interpolating the neighborhood features, and there is no learnable weight for the interpolation. This limits the description ability of the feature.

We make a simple modification by adding a set-abstraction layer before each feature propagation layer to overcome this limitation. As shown in Fig.\ref{fig:framework}(b), we create new features for up-sampled points by grouping their neighborhood in the old point clouds and passing it to a multilayer perceptron (MLP). After that, we concatenate the new features with the interpolated features and feed them to the MLP like standard feature propagation layers described in \cite{Qi2017_NIPS}. This modification allows the set-abstraction layers to provide more neighbor information and learn-able weights for interpolation. 

We propose a 2D CNN  network with a U-Net structure for image feature learning. The encoder is a ResNet having four ResNet layers\cite{He2016_CVPR}, while the decoder is an inverse of the encoder that replaces the first convolution layer with the deconvolution layer to up-sample the feature. Convolutional neural networks (CNNs) are inherently unable to handle non-trivial geometric transformations due to the fixed geometric structures in their building modules\cite{Bronstein2017_ispm}. The PointNet++, on the other hand, is more flexible for modeling geometric transformations and has a broader receptive field view. To make the 2D model more flexible, we replace the second convolutional layers of the ResNet block in the encoder with deformable convolutional layers\cite{Dai2017_ICCV}. The same modification is made in the decoder model.
\begin{figure}[t]
  \centering
  \includegraphics[width=0.48\textwidth]{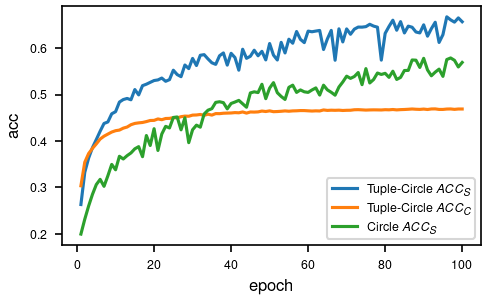}
  \caption{ Validation accuracy of Tuple-Circle Loss (blue) vs. Circle Loss (green) during the training Process}
  \label{fig:circlevstuple}
\end{figure}
\begin{center} 
\begin{figure*}[t]
\setlength{\tabcolsep}{1pt}
\begin{tabular}{ccc} 
\includegraphics[width=0.33\textwidth]{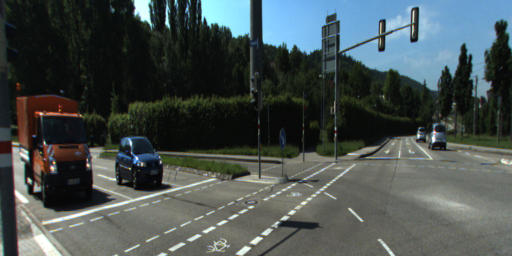}&
\includegraphics[width=0.33\textwidth]{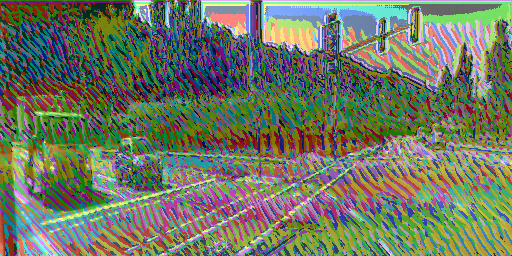}&
\includegraphics[width=0.33\textwidth]{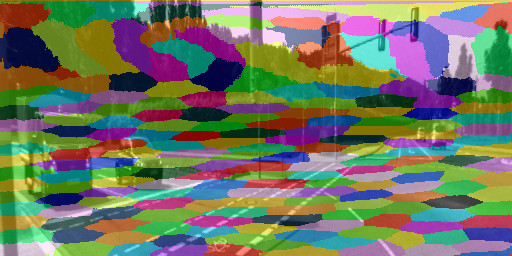}\\
(a)&(b)&(c)\\
\includegraphics[width=0.33\textwidth]{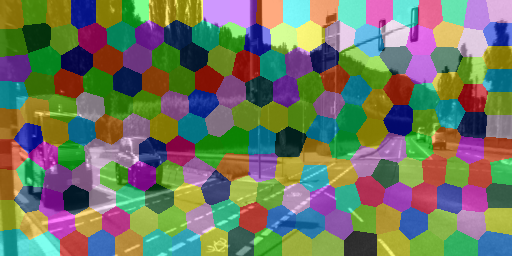}&
\includegraphics[width=0.33\textwidth]{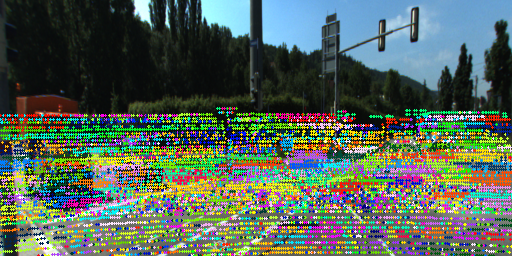}&
\includegraphics[width=0.33\textwidth]{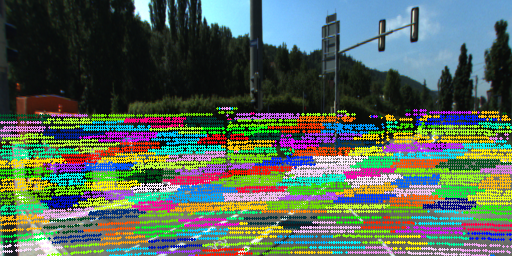}\\
(d)&(e)&(f)\\
\end{tabular} 
\caption{\textbf{Feature KMean Clustering Visualization}: (a) Original image (b) Full feature clustering results of image; (c) Shared feature clustering results of image; (d) Pixel coordinate clustering results; (e)Full feature clustering results of point cloud; (f)  Shared feature clustering results of point cloud;} 
\label{fig:label_viz}
\end{figure*} 
\end{center}  
\section{EXPERIMENTAL EVALUATION}
\subsection{Dataset}
In order to verify our approach, the evaluation dataset is required to have the image and point cloud pairs with known pixel-point level correspondence (known sensor calibration). Datasets meeting this requirement exist in papers, such as \cite{Liao2021_ARXIV, Jiang2021_RELLIS3D}. Other works like \cite{Wang2022_p2net, Pham2019_LCD} construct datasets in order to have dense correspondence between point and pixel. However, not to constraint to our model to dense point cloud and show the generality of our method, we use camera image and LiDAR scan pairs sequences from the KITTI360 dataset\cite{Liao2021_ARXIV}. We trained our models on sequences 0, 2, 4, 5, 6, 7, and 9 of KITTI360 and performed validation on sequence 3. During training, the image sequences were randomly cropped to $256\times512$, and the LiDAR point cloud was down-sampled to $10000$ points.
\subsection{Implementation}
A lambda workstation with two NVIDIA Titan RTX was used for training, and, Pytorch library \cite{Paszke2019_NIPS} was used to implement the models. The learning rate was initialized at 0.01 and decayed every epoch by 0.985 for 100 epochs. Finally, we used AdamW optimizer to optimize the models\cite{Ilya2019_ICLR}, and the total size of the feature vector was 256 for all the models.
\subsection{Evaluation}
Since our goal is to find the unified cross-modal features between two modalities and not to detect key points for matching, we evaluate our features by randomly sampling $500$ points from two modalities and calculating the percentage of correct matches. For inter-modal matching, we use full features $[x_{sh},x_{pr}]$ to compute cosine similarity. For cross-modal matching, we only use the shared part features $x_{sh}$ to perform matching. In the following part of the paper, $ACC_I$ and $ACC_P$ denote the inter-modal matching accuracy of image and point cloud, respectively. $ACC_C$ denote the cross-modality matching accuracy using full feature vectors $[x_{sh},x_{pr}]$, and $ACC_S$ denotes the cross-modal matching accuracy using shared feature vectors $x_{sh}$. 
\subsubsection{Tuple-Circle Loss vs. Circle Loss}
\begin{figure}[h]
   \centering
    \includegraphics[width=0.49\textwidth]{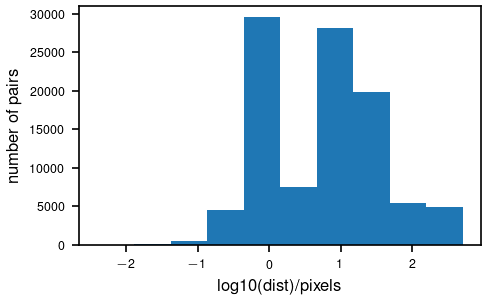}
    \caption{Histogram of distance between the mismatched cross-modal pairs in $log_{10}$ scale. x-axis denotes the distance between the mismatched cross-modal pairs in $log_{10}$ scale; y-axis denotes the number of pairs.}
    \label{fig:misdist_hist}
\end{figure}

In our preliminary test, other widely used contrastive loss functions \cite{Chopra2005_CVPR, Gutmann2010_PMLR, Van2018_arXiv} can easily allow the two models for different modalities to learn good inter-modal features while training simultaneously. However, two models cannot or only slowly converge to learn cross-modal features. Thus, we compare the proposed Tuple-Circle loss function with the Circle loss function \cite{Sun2020_Circleloss, Wang2022_p2net}. Fig.\ref{fig:circlevstuple} shows the accuracy changes during training. There are three lines in Fig.\ref{fig:circlevstuple} which denote the $ACC_S$ of Tuple-Circle loss (blue) and Circle loss (green), and $ACC_C$ of Tuple-Circle loss. We can see that the Tuple-Circle loss function converges faster than the Circle loss function and converges to better results. Another interesting observation of Tuple-Circle loss is that the full features can also be used to distinguish across modalities and show better performance at the early stage. However, after some epochs, the full features cannot get better results and are outperformed by the shared features. The Table \ref{tbl:model_cmp} shows the comparison of different model settings. Rows $1$ and $2$ show that using Tuple-Circle loss results in better feature learning for inter and cross-modality matching. However, cross-modal matching accuracy is much lower than inter-matching accuracy. Fig.\ref{fig:misdist_hist} shows that the distance between more than 40\% of mismatched cross-modal pairs is less than 1.5 pixels after projecting the LiDAR point cloud on the image.
\subsubsection{Models Comparison}
We also test the performance of different model setups described in setion\ref{models} on feature learning. Table \ref{tbl:model_cmp} shows the comparison of different model settings; DCN denotes the use of Deformable convolution layers in the 2D model, and ASFP denotes using the set abstraction forward propagation. The results show that ASFP layers help to learn better point cloud features and cross-modality features (highlighted in bold in Table.\ref{tbl:model_cmp}). According to \cite{Wang2022_p2net}, a broader reception field help to learn inter and cross features, and a Deformable convolution layer can provide a larger and more flexible reception field \cite{Dai2017_ICCV}. However, the DCN layer did not help us learn cross features from our experiment results.
{\small  
\begin{table}[!ht]
    \centering
    \caption{Comparison of Different Model Settings}
    \begin{tabular}{c|c|c|c|c|c|c}
    \hline
      Tuple & DCN & ASFP& $ACC_{I}$ & $ACC_{P}$ & $ACC_{C}$  & $ACC_{S}$\\ \hline
     \xmark  & \cmark & \cmark&96.1\% & 91.5\% & - & 57.9\%\\ \hline
    \cmark & \cmark & \cmark& \textbf{99.9\%} & \textbf{97.3\%}  & \textbf{46.9\%} & \textbf{66.7\%}\\ \hline
     \cmark  & \xmark & \xmark&99.9\% & 78.6\% & 45.3\% & 63.1\%\\ \hline
     \cmark & \xmark & \cmark&99.9\% & 97.4\%  & 47.0\% & 67.4\%\\ \hline
    \end{tabular}
    \label{tbl:model_cmp}
\end{table}}
{\small  
\begin{table}[!ht]
    \centering
    \caption{Comparison of Different Shared Feature Size}
    \begin{tabular}{c|c|c|c|c}
    \hline
      Shared Size & $ACC_{I}$ & $ACC_{P}$ & $ACC_{C}$  & $ACC_{S}$\\ \hline
      64 &99.9\% & 97.5\% & 45.0\% &67.9\%\\ \hline
      128 & 99.9\% & 97.4\%  & 47.0\% & 67.4\%\\ \hline
      192 & 99.9\% & 97.3\% & 46.3\% & 68.0\%\\ \hline
    \end{tabular}
    \label{tbl:share_cmp}
\end{table}}
\subsubsection{Shared Features Size Comparison}
We also studied the effect of different sizes of shared features on the performance of feature learning. The results are presented in Table.\ref{tbl:share_cmp}. Interestingly, the inter-matching results improve after introducing the Tuple-Circle loss. However, the cross-modality matching results did not change much with the change in the size of the shared features. We hypothesize that the learned shared information cross-modality does not have very high entropy.

\subsection{Visualization}
What do the two models learn across different modalities? In order to answer this question and visualize the features, we use Cosine-KMean Clustering \cite{Arindam2005_JMLR} method. We first use the full features $x_{f}=\left[x_{sh},x_{pr}\right]$  as input to the Cosine-KMean Clustering method and try to classify these features in 200 clusters. The results are shown in Figs.\ref{fig:label_viz}(b). We also perform the same clustering on full features of the image's corresponding point cloud and show the result in  Figs.\ref{fig:label_viz}(e). As shown, we cannot find many similarities between Figs.\ref{fig:label_viz}(b) and (e). Fig.\ref{fig:label_viz}(e) looks more like noise due to the low resolution of the point cloud. However, we can find a clear pattern in Figs.\ref{fig:label_viz}(b). The pattern shows more texture properties than geometric properties, which have been shown in \cite{Robert2018_CoRR, Leng2019_access}. For shared features $x_{sh}$, we first concatenate features from both modalities as a dataset and try to classify all 200 clusters by using the Cosine-KMean Clustering method. And then, we plot clustering results of pixel and point in Figs.\ref{fig:label_viz}(b) and (d). As we can see, both image and point cloud clusters overlap. Moreover, the clustering results show the position relative properties. In Fig.Fig.\ref{fig:label_viz}(d), we show an image pixel position clustering results. We use all the pixel positions $(row, column)$ in an image as a dataset and perform normal KMean clustering on the dataset. In this way, we can get an image like Fig.\ref{fig:label_viz}(d), and we can see this result is close to the Voronoi diagram. Comparing Fig.\ref{fig:label_viz}(d) and (c), we can observe some similarities between the two figures. Fig.\ref{fig:label_viz} (c) shows that the shared features may encode 2D position information, which is shown in Fig.\ref{fig:label_viz}(d) and the 3D depth information of the corresponding point cloud which is shown in Fig.\ref{fig:label_viz}(f). 

\begin{center} 
\begin{figure}[h]
\begin{tabular}{c} 
\includegraphics[width=0.45\textwidth]{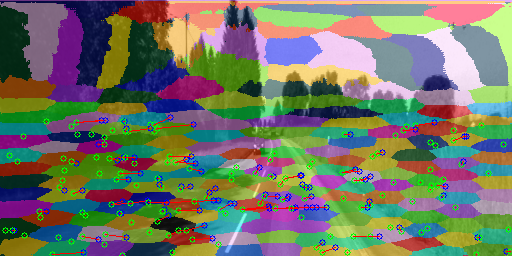}\\(a)\\
\includegraphics[width=0.45\textwidth]{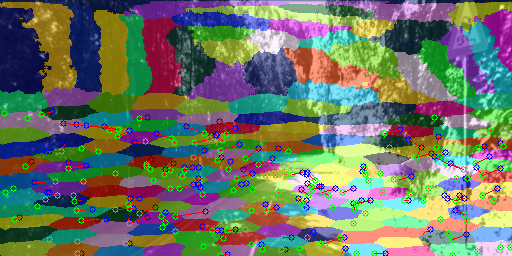}\\(b)\\
\end{tabular} 
\caption{\textbf{Cross-modality Matching Visualization}: Green and Blue circles represent points from the point cloud and image individually; Red lines denote the mismatching; (Note single green due to the small distance between two mismatched points.)} 
\label{fig:match_viz}
\end{figure}
\end{center}

We also analyze the error matching across different modalities based on the clustering results. Figs.\ref{fig:match_viz} (a) and (b) show part of the mismatching. From the visualization, it can be observed that most mismatching happened within the same clusters or on the border of two clusters. 
\section{CONCLUSIONS}
This paper proposes a Tuple-Circle loss function for cross-modality feature learning to learn common features across different sensor information modalities. Our results show that the proposed Tuple-Circle loss allows faster and better convergence of the model. Furthermore, we develop and present a variant of PointNet++ architecture for point cloud to achieve better inter-modal and cross-modal matching. We develop a variant of U-Net CNN architecture for image feature learning and study the effectiveness of the Deformable convolutional layers for our evaluation settings. By utilizing the Cosine-Kmean clustering method, we present the visualizations of the learned features and show that our method allows the two models of different modalities to learn geometrically meaningful common features.  





\bibliographystyle{IEEEtran}
\bibliography{IEEEabrv,references}

\end{document}